\documentclass[11pt,letterpaper]{article}
\usepackage{cogsys}
\usepackage[T1]{fontenc}
\usepackage{times}
\usepackage[pdftex]{graphicx} 

\graphicspath{ {pic/} }
\usepackage{amsmath}
\usepackage{booktabs}
\usepackage{subcaption}
\usepackage{multirow}
\usepackage{tabularx}
\usepackage{rotating}
\usepackage[flushleft]{threeparttable}
\usepackage[]{hyperref}
\usepackage{enumitem}
\hypersetup{
    pdftitle={Automatic Item Generation of Figural Analogy Problems: A Review and Outlook},
    bookmarksnumbered=true,     
    bookmarksopen=true,         
    bookmarksopenlevel=1,       
    colorlinks=true,  
    citecolor=[rgb]{0, 0.5, 0.5},
    linkcolor=[rgb]{0, 0.5, 0.5},
    pdfstartview=FitH,
    pdfpagelayout=SinglePage
}

\usepackage{natbib}
\cogsysheading{11}{2021}{1--**}{9/2021}{11/2021}

\ShortHeadings{Automatic Item Generation of Figural Analogy Problems}{Y.\ Yang, D. \ Sanyal, J. \ Michelson, J. \ Ainooson, and M. \ Kunda}

\begin{document} 

\title{Automatic Item Generation of Figural Analogy Problems:\\ A Review and Outlook}
\author{Yuan Yang}{yuan.yang@vanderbilt.edu}
\author{Deepayan Sanyal}{deepayan.sanyal@vanderbilt.edu}
\author{Joel Michelson}{joel.p.michelson@vanderbilt.edu}
\author{James Ainooson}{james.ainooson@vanderbilt.edu}
\author{Maithilee Kunda}{mkunda @vanderbilt.edu}
\address{Electrical Engineering and Computer Science, Vanderbilt University, Nashville, TN 37235 USA}
\vskip 0.2in
 
\begin{abstract}
Figural analogy problems have long been a widely used format in human intelligence tests.  In the past four decades, more and more research has investigated automatic item generation for figural analogy problems, i.e., algorithmic approaches for systematically and automatically creating such problems. In cognitive science and psychometrics, this research can deepen our understandings of human analogical ability and psychometric properties of figural analogies. With the recent development of data-driven AI models for reasoning about figural analogies, the territory of automatic item generation of figural analogies has further expanded. This expansion brings new challenges as well as opportunities, which demand reflection on previous item generation research and planning future studies. This paper reviews the important works of automatic item generation of figural analogies for both human intelligence tests and data-driven AI models. From an interdisciplinary perspective, the principles and technical details of these works are analyzed and compared, and desiderata for future research are suggested.

\end{abstract}

\section{Introduction}

\begin{figure}[ht]
    \centering
    \begin{subfigure}{0.35\textwidth}
        \centering
        \includegraphics[width=0.97\textwidth]{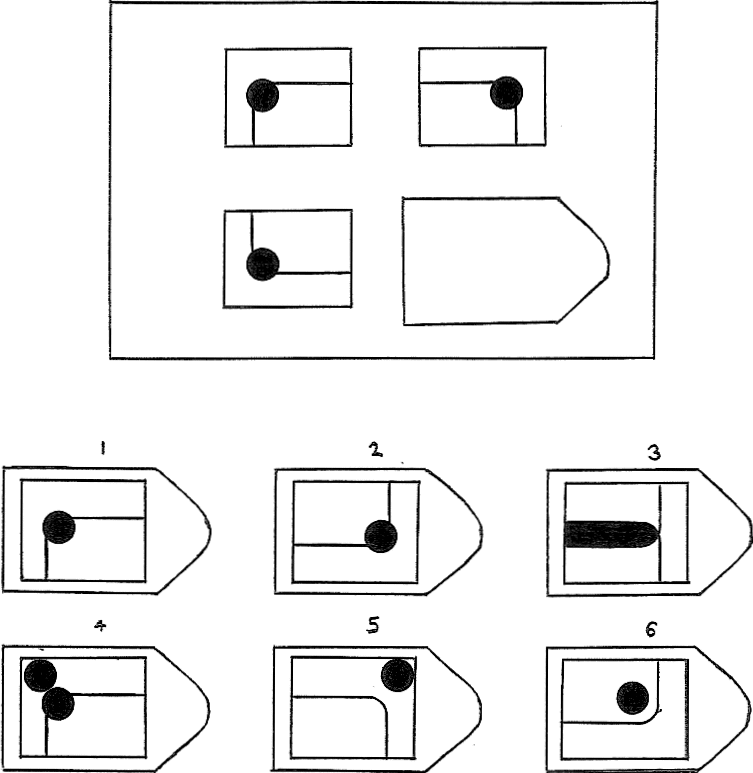} 
        \caption{2$\times$2 Progressive Matrix \citep{kunda2013computational}}
        \label{fig:22rpm}
    \end{subfigure}
    \hspace{0.005\textwidth}
    \begin{subfigure}{0.35\textwidth}
        \centering
        \includegraphics[width=\textwidth]{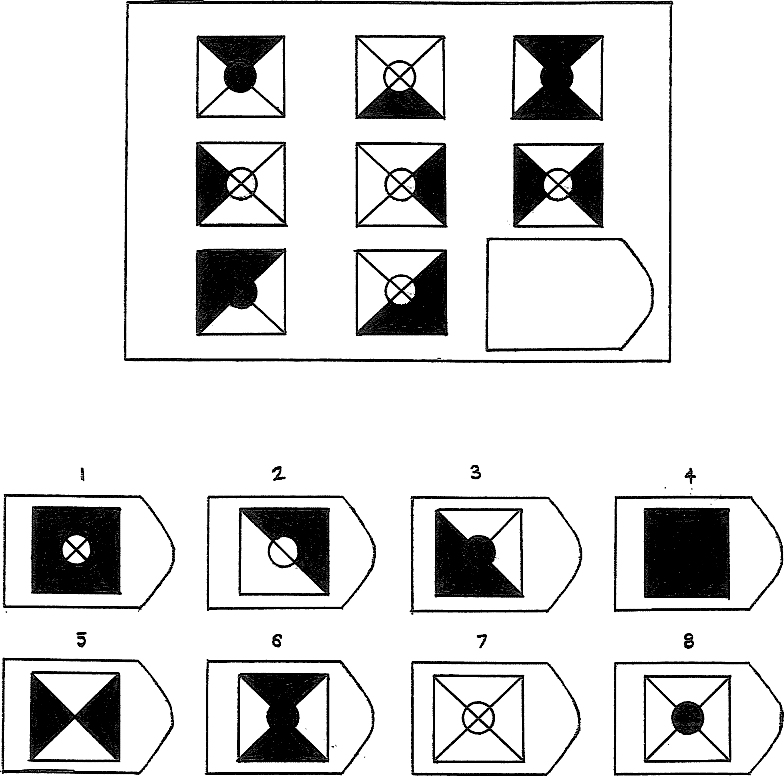}
        \caption{3$\times$3 Progressive Matrix \citep{kunda2013computational}}
        \label{fig:33rpm}
    \end{subfigure}
    \vspace{0.05\textwidth}
    
    \begin{subfigure}{0.35\textwidth}
        \centering
        \includegraphics[width=\textwidth]{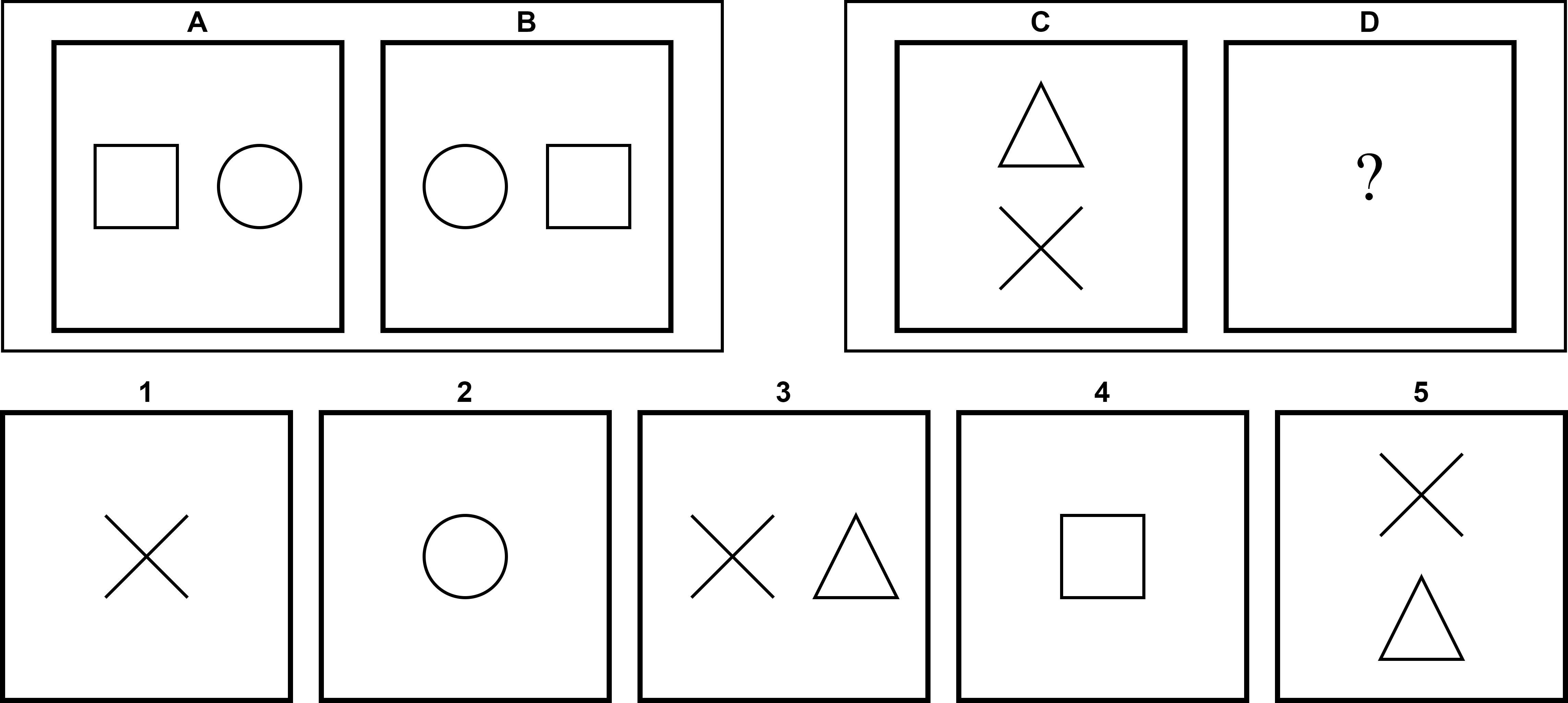}
        \caption{2\textbf{$+$}2 Geometric Analogy \citep{lovett2009solving}}
        \label{fig:22gap}
    \end{subfigure}
    \hspace{0.005\textwidth}
    \begin{subfigure}{0.35\textwidth}
        \centering
        \includegraphics[width=0.9\textwidth]{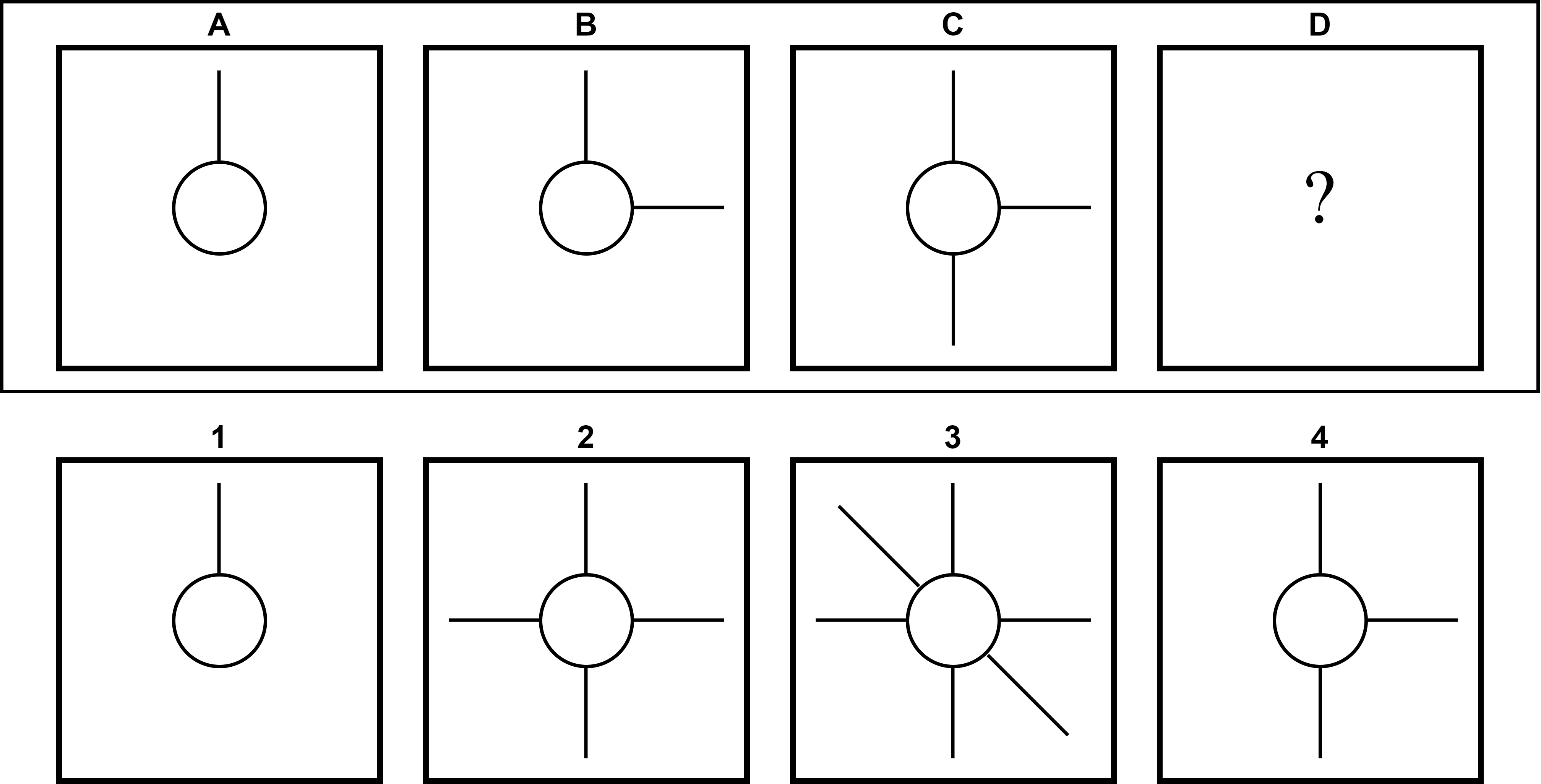}
        \caption{4$\times$1 Figural Series \\ \citep{sekh2020can}}
        \label{fig:41fs}
    \end{subfigure}
\caption{Examples of figural analogy problems.}
\label{fig:figural-analogies}
\end{figure}

Figural analogy (FA) tasks (sometimes called geometric analogy, matrix reasoning, etc.) have long occupied a very central position among human intelligence tests \citep{snow1984topography}, and have also had a long history of research in artificial intelligence (AI).
 An FA is usually represented as a 2D array of figures. 
Figure \ref{fig:figural-analogies} shows 4 common formats of figural analogy problems (FAPs). Figure \ref{fig:22rpm} and \ref{fig:33rpm} exemplify the Raven's Progressive Matrices (RPM), which is widely used as the single intelligence test whose score is most representative of the general intellectual ability. Figure \ref{fig:22gap} shows another FA task --- proportional geometric analogies --- that were attempted by the first computational solver of FAPs \citep{evans1964program}. 
Figure \ref{fig:41fs} shows another common FA task in intelligence tests --- figural series.

Many computational models for solving analogy problems have been constructed in different disciplines, such as psychometrics, neuroscience, cognitive science, and artificial intelligence. Although these models are different in their purposes and theoretical foundations depending on what disciplines they belong to, they do share a great demand for high-quality analogy problems that should be as diverse and representative as possible in their particular analogy domain. In particular, since the data-driven approaches in AI joined the game, this demand has soared up exponentially, and instantly dwarfed the reserve of analogy items and the productivity of human item writers of analogy problems. 

The purpose of this paper is to seek possible solutions to this supply-demand gap in a different line of research --- automatic item generation (AIG). AIG refers to the approach of using computer algorithms to automatically create testing items. It has been adopted and studied for decades in areas such as psychology, psychometrics, cognitive science, and education. It generates a wide range of testing items from domain-general tests, such as human IQ tests \citep{hornke1986rule}, to domain-specific tests, such as medical license tests \citep{gierl2012using}. AIG was initially introduced to address the increased demand for testing items in the following test settings:
\begin{itemize}[nolistsep,noitemsep]
    \item Large-scale ability testing, for example, repeated tests in academic settings and longitudinal experiments, where many parallel forms are needed due to the retest effect.
    \item Adaptive testing, in which the next items are determined by the responses to previous items, which is a more efficient and reliable testing form, but also requires larger item banks.
    \item Computer-based and internet-based testing, which makes standardized tests more accessible to the public and brings the exposure control issue to a new level.
\end{itemize}

In addition, recent work on data-driven FA approaches in AI creates another type of demand --- datasets to fit and evaluate AI models. More importantly, as the usage of testing items in AI may be vastly different from that in human testing, this presents a new research question about how AIG should evolve to properly serve all the research purposes in these different usages --- constructing AI models, understanding human cognition, and evaluating both AI and human performance.

In this paper, we will first review the methods of automatically generating FAPs for human intelligence tests. Then, we switch to the works of automatically generating FAPs for fitting and evaluating data-driven AI models. However, our review is by no means exhaustive, and we select the most representative ones to analyze. At last, we conclude with reflections on the reviewed works and suggest possible future directions for AIG of FAPs.

\begin{sidewaystable}
\small
\centering
\caption{The technical details of generator programs in the reviewed works.}
\label{tab:tech-details}

\begin{threeparttable}

\begin{tabularx}{\textwidth}{X X X X X X X X X X X X X}
\toprule
    \multirow{2}{*}{Model} & \multirow{2}{1cm}{Format}   & \multirow{2}{2cm}{Analogical Direction\tnote{a}} & \multicolumn{6}{c}{Geometric Element} & \multicolumn{2}{c}{Rules} & \multirow{2}{2cm}{Per. Org.\tnote{d}} &  \multirow{2}{1cm}{Answer Set\tnote{e}} \\
    \cmidrule(lr){4-9} \cmidrule(lr){10-11}
    & & & Type\tnote{b} & Size & Color & Angle & Number & Position & Type\tnote{c} & Number & & \\
\midrule
    Rule-Based & 3$\times$3 & R, C, O & - & - & - & - & 1-2 & - & I, S, P, D & 1-2 & S, I, E & - \\
    \hline
    Cognitive & 3$\times$3 & R, C, O & - & - & - & - & - & - & I, S, P, D & 1-2 & O, F, D & 8 \\
    \hline 
    GeomGen & 3$\times$3 & R, C & S, C,T, D2, H, R & -  & - & - & - & 3$\times$3 grid & I, S, P, N & - & C, N & 5+1\\
    \hline
    Sandia & 3$\times$3 & R, C, M, S, O & O, R, T, D1, T1, T2 & 5 sizes & 5 colors & 5 angles & 1-6 & center & S, P & 1-3 & center, overlay & 8 \\
    \hline
    CSP & 3$\times$3 & R, C & - & - & - & - & - & - & S, P, D & 1-7 & - & 8 \\
    \hline
    IMak & 2$\times$2 & O & C, L, D0, T1 & 1 size & 1 color & 8 angles & 3 & - & P & 1-4 & E & 8+2 \\
    \hline
    PGM-shape & 3$\times$3 & R, C & C, T, S, P, H & 10 sizes & 10 colors & - & 0-9 & 3$\times$3 grid & S, P, D & 1-4 & 3$\times$3 grid & 8 \\
    \hline
    PGM-line & 3$\times$3 & R, C & L, D1, C & 1 size & 10 colors & 1 angle & 1-5 & center & S, P, D & 1-4 & center, overlay & 8 \\
    \hline
    RAVEN & 3$\times$3 & R & T, S, P, H, C & 6 sizes & 10 colors & 8 angles & 1-9 & fixed in configurations & I, S, P, A, D & 4 & 7 configurations & 8 \\
\bottomrule
\end{tabularx}

\begin{tablenotes}
    \item[a] R=row, C=column, M=main-diagonal, S=secondary-diagonal, O=R+C (see \citep{matzen2010recreating}).
    \item[b] S=square, C=circle, R=rectangles, T=triangles, H=hexagons, O=oval, L=line, D0=dot D1=diamond, T1=trapezoid, T2=``T'', D2=deltoid, P=pentagon..
    \item[c] I=identify, S=set/logical operations, P=progression, D=Distribution of 2 or 3 values, N=neighborhood, A = arithmetic (see \citep{carpenter1990one, arendasy2002geomgen}).
    \item[d] S=separation, I=integration, E=embedding, C=classical, N=normal, O=overlay, F=fusion, D=distortion (see \citep{hornke1986rule, embretson1998cognitive, primi2001complexity}).
    \item[e] +1=``none of the above'', +2=``none of the above'' + ``I don't know''.
\end{tablenotes}

\end{threeparttable}
\end{sidewaystable}

\section{Automatically Generating Figural Analogy Problems for Human Intelligence Tests}
 
Human intelligence tests consist of items carefully handcrafted by psychologists and psychometricians. However, item writing is more of an art than a science due to its idiosyncrasy and inefficiency. Handcrafted items have to go through iterations of evaluation and calibrating for good psychometric properties before being included in the final item bank. The attrition rate could be up to 50\% \citep{embretson2004measuring}. A variety of efforts in AIG have been made to free item writers from the onerousness of item writing. In this section, we discuss the important works of automatically generating FAPs for intelligence tests, each of which is described in a subsection, whose title is the name of the work followed by a keyword of its most outstanding characteristic. The technical details of the works are summarized in Table \ref{tab:tech-details}.


\subsection{Rule-Based Item Construction --- Human-Based AIG}

\citet{hornke1986rule} conducted one of the earliest studies, if not the earliest, on AIG.  
Although they did not literally automatically generate items, what they achieved was already very close to today's AIG. They created a standard procedure for item generation, hired university students to manually execute this procedure, and created 648 3$\times$3 FAPs. Each step in this procedure had finite clearly defined options so that human idiosyncrasy was kept at the minimum level. To the best of our knowledge, although the diversity and complexity of these items were not comparable to expert designs, no one had ever ``automatically'' created so many items.

\begin{figure}[ht]
    \centering
    \includegraphics[width=\textwidth]{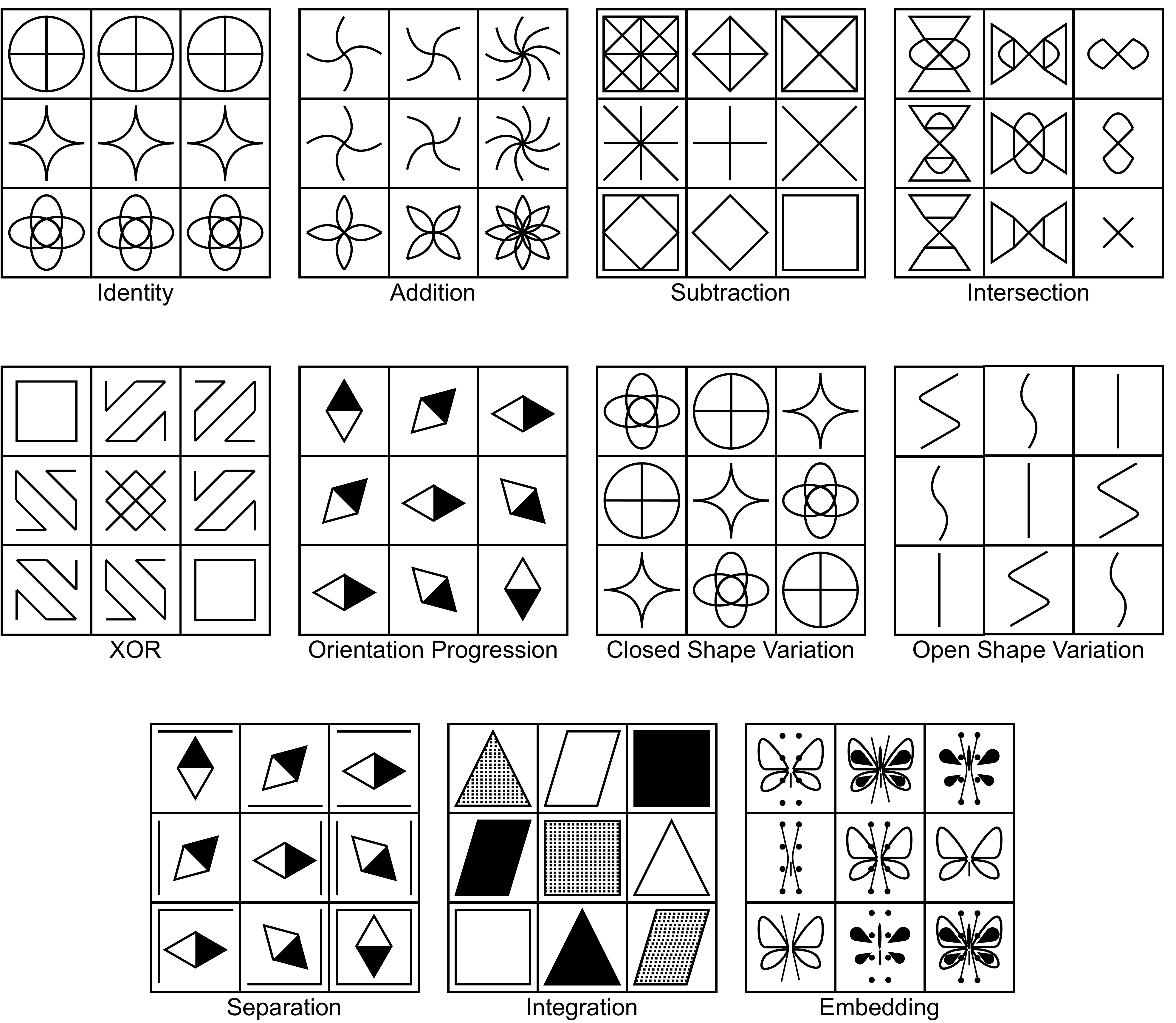}
    \caption{Example items created by following \citeauthor{hornke1986rule}'s AIG procedure.}
    \label{fig:hb}
\end{figure}

\citeauthor{hornke1986rule} considered the item writing task as the reverse of the solving process, which could be decomposed into three types of cognitive operations. The three types addressed three common independent dimensions of solving or creating an FAP item. To generate items, \citeauthor{hornke1986rule} thus designed a procedure that sequentially made choices on these three dimensions by selecting from finite sets of options:
\begin{itemize}[nolistsep,noitemsep]
    \item Variation rules of geometric elements: eight options are provided (see the first 8 matrices in Figure \ref{fig:hb} for examples): identity, addition, subtraction, intersection, exclusive union (or symmetric difference), progression, variation of open/closed gestalts (i.e. distribution of three hollow/solid shapes). 
    \item Analogical directions: a variation rule proceeds in row or column direction.
    \item Perceptual organizations: this dimension addresses how multiple variation rules are combined into a single item. Three options are provided (see the last 3 matrices in Figure \ref{fig:hb} for examples): separation, integration, and embedding. Separation means that separate geometric elements are used for different variation rule; integration means that different attributes of a single geometric element are used for different variation rules; and embedding means that different parts of a single geometric element are used for different variation rules.
\end{itemize}

In their experiment, student item writers were given a set of geometric elements (e.g. differently sized squares and triangles) and instructed to create items by jointly sampling the 3 dimensions and geometric elements from the given set. The students were told to create each items by combining at most two variation rules. Therefore, the resulting item bank contained only 1-rule and 2-rule items. Human experiments on this item bank showed that the cognitive operations corresponding to these 3 dimensions explained roughly 40\% of the item difficulty. As for the unexplained 60\%, other early studies \citep{mulholland1980components} indicated that the numbers of elements and rules were also major sources of difficulty. 

\subsection{Cognitive Design System Approach --- Cognitive Modeling}

\citet{embretson1995role, embretson1998cognitive, embretson2004measuring} introduced the Cognitive Design System Approach.  
Based on the information processing theory of \citet{carpenter1990one}, an item bank of FAPs was generated as a demonstration.
The Cognitive Design System Approach starts with the cognitive modeling of the solving process of an existing ability test. \citeauthor{embretson2004measuring} pointed out that 
the cognitive model in \citep{carpenter1990one}, which many following AIG works were based upon, did not include perceptual encoding or decision processes in the solving process.
Hence \citeauthor{embretson2004measuring} incorporated three extra binary perceptual stimulus features --- object overlay, object fusion, and object distortion --- in the Cognitive Design System Approach, which represent three different types of mental decomposition of the complete gestalt into its basic parts. Essentially, object overlay and fusion are similar to separation and embedding in Figure \ref{fig:hb}, while object distortion refers to perceptually altering the shape of corresponding elements (e.g. bending, twisting, stretching, etc.). A software, ITEMGEN, was developed based on this approach, but the software is not easily accessible now.

Once the cognitive models are determined, the stimulus features are accordingly determined. The Cognitive Design System Approach then employs psychometric models to estimate item properties (such as item difficulty and item discrimination), formulated as parameterized functions of the stimulus features. The function parameters are initially set by fitting the psychometric models to human data on the existing tests. Thereafter, the psychometric properties of newly generated items (by manipulating the stimulus features) could be predicted by these functions. The prediction and empirical analysis of the newly generated items are compared to further calibrate the parameters until the functions are sufficiently predictive. The approach could be integrated into an adaptive testing system to replace a fixed item bank and generate items of expected properties in real-time.


\subsection{MatrixDeveloper --- 4$\times$4 Matrices}

MatrixDeveloper \citep{hofer2004matrixdeveloper} is an unpublished software that generates FAPs. 
It has been used in several studies of psychometric properties of computer-generated FAPs  \citep{freund2008explaining,freund2011get, freund2011retest, freund2011wants}. According to the limited description in these studies, the MatrixDeveloper is similar to the Cognitive Design System Approach in terms of variation rules (i.e. the five rules summarized in \citep{carpenter1990one}) and perceptual organizations (i.e. overlap, fusion, and distortion). The difference is that it generates 4$\times$4 matrix items. Therefore, it could accommodate more variation rules compared to 3$\times$3 or 2$\times$2 matrices so that the differential effects of types of variation rules could be better studied.

\subsection{GeomGen --- Perceptual Organization}

The early cognitive modelings of solving hand-crafted FAPs tended to characterize the items by the numbers of elements and rules and types of rules, for example, \citep{mulholland1980components,bethell1984adaptive,carpenter1990one}. 
However, unlike solving the existing items, creating new items requires us to consider at least one more factor --- perceptual organization \citep{primi2001complexity}. It tells how geometric attributes, geometric elements and rules are perceptually integrated to render the item image. For example, the third dimension in the procedure of \citep{hornke1986rule} is a specific way to deal with perceptual organization. More generally, perceptual organization involves the Gestalt grouping/mapping of elements using Gestalt principles such as proximity, similarity, and continuity. This factor is less clearly defined and less systematically investigated (and whether it is a unidimensional factor is even questionable). But, to create new items, one has to adopt some formalized ways to manipulate perceptual organization.


\citep{arendasy2002geomgen, arendasy2005effect} proposed a generator program --- GeomGen --- that adopted a binary perceptual organization, which was reused and extended in many following works.
The perceptual organization in GeomGen was classified into classical view and normal view. In classical views, the appearance of geometric elements changes while numbers and positions of them remain constant. In normal views, numbers and positions of elements change while the appearance remain constant. The appearance changes, such as in size and aspect ratio or distortions, should preserve the Gestalt grouping/mapping of visual similarity. Therefore, in both classical and normal views, the type (i.e. geometric shape) of elements are set to be constant in GeomGen. 


The taxonomy of perceptual organization in GeomGen is only a specific way to define perceptual organization but by no means the unique way. For example,  \citet{primi2001complexity} proposed another important taxonomy --- harmonic and nonharmonic, which, together with GeomGen taxonomy, forms a better description of perceptual organization, which is adopted in many following works.  

\begin{figure}[ht]
    \centering
    \includegraphics[width=0.8\textwidth]{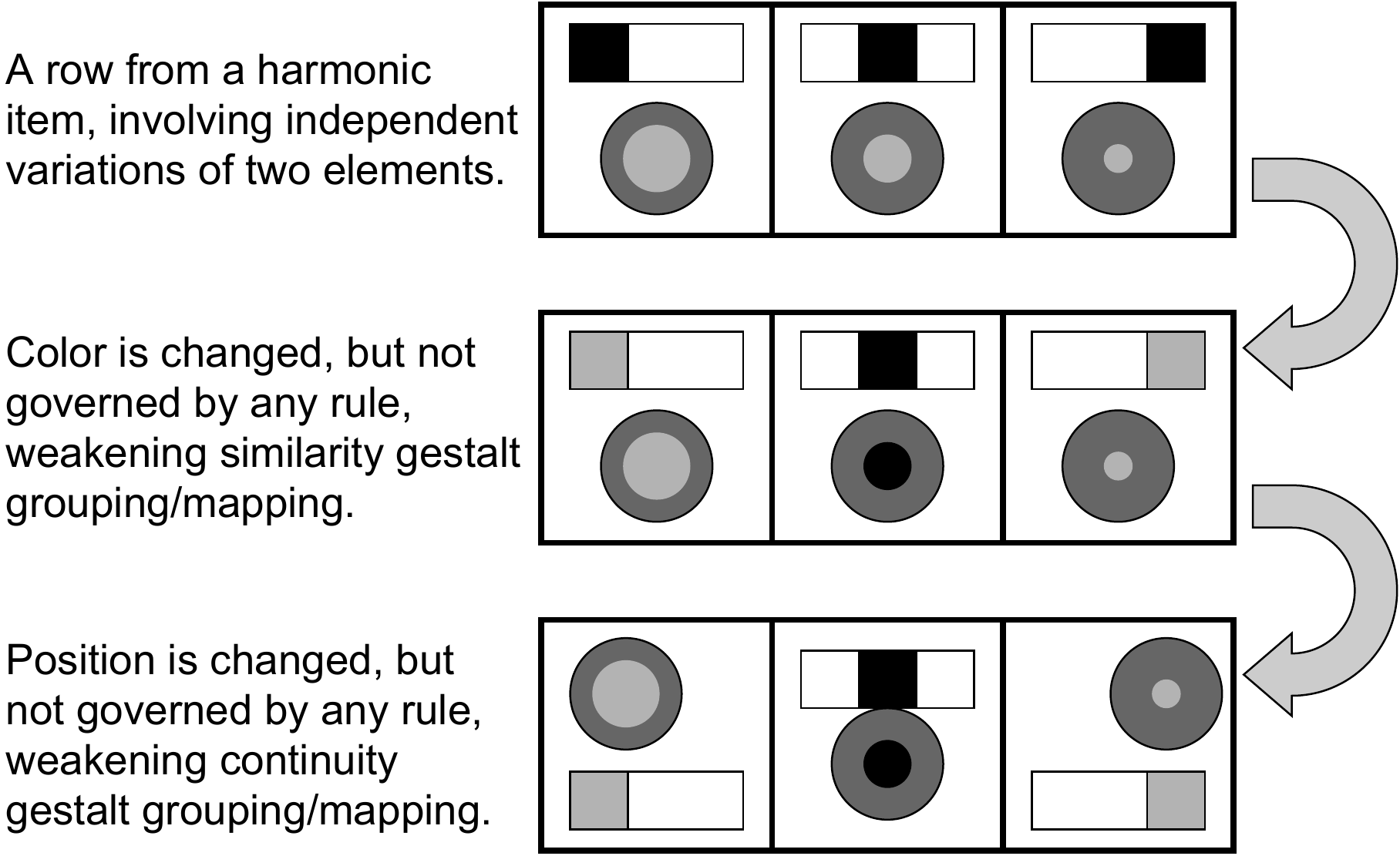}
    \caption{An example of deriving nonharmonic items from harmonic items.}
    \label{fig:harmonic}
\end{figure}

\citet{primi2001complexity} classified perceptual organizations into harmonic and nonharmonic ones--- ``visually harmonic items display perceptual and conceptual combinations that represent congruent relationships between elements, whereas nonharmonic items tend to portray competitive or conflicting combinations between visual and conceptual aspects that must be dealt with in reaching a solution.'' \citet{primi2001complexity} mentioned that, in the practice of AIG, the nonharmonic items could be derived from the harmonic ones by manipulating the attributes of elements to cause misleading Gestalt groupings, as shown in Figure \ref{fig:harmonic}. The correct Gestalt grouping/mapping (i.e. element correspondences) are obvious in harmonic items, whereas nonharmonic items requires extra cognitive effort to resolve the conflict between competing gestalt groupings and mappings.


\subsection{Sandia Matrix Generation Software --- High-Fidelity SPM Generator}

Based on the Raven's Standard Progressive Matrices (SPM), \citet{matzen2010recreating} presented the Sandia Matrix Generator that could ``recreate'' the 3$\times$3 SPM items with high fidelity.

\citet{matzen2010recreating} identified two basic types of 3$\times$3 items in SPM--- the element transformation and the logic problems. An element transformation refers to a progressive variation of a certain attribute of the element. There could be multiple variations in different directions, for example, a color variation in the row direction and a size variation in the column direction.
The attributes considered for transformation problems are shape, shading, orientation, size, and number, each of which takes values from an ordered categorical domain. The logic problems involve operations such as addition/subtraction, conjunction (AND), disjunction (OR), or exclusive disjunction (XOR) of elements along rows. The elements in logic problems are overlaid with one another. Each generated item can be either a transformation one or a logic one, but not both.

The Sandia Matrix Generator generates answer choices in ways that are common among the original SPM problems. An incorrect answer choice could be (a) an entry in the matrix, (b) a random transformation of an entry in the matrix, (c) a random transformation of the correct answer, (d) a random transformation of an incorrect answer, (e) a combination of features sampled from the matrix, and (e) a combination of novel features that did not appear in the matrix. 

The item difficulty was studied through an item bank of 840 generated items. 
Human experiment data showed that the generated items and the original SPM had generally very similar item difficulty. 
In particular, the data further showed that the item difficulty was mostly affected by the number of rules, analogical directions, and problem types.  

\subsection{CSP Generator --- First-Order Logic Representation}


\citet{wang2015automatic} made an effort to represent 3$\times$3 RPM items in a more formal manner --- the first-order logic --- and turned the AIG problem into a constraint satisfaction problem (CSP) by formulating the ``validity'' of RPM items into a set of first-order logic propositions.

\begin{figure}[ht]
    \centering
    \includegraphics[width=.8\textwidth]{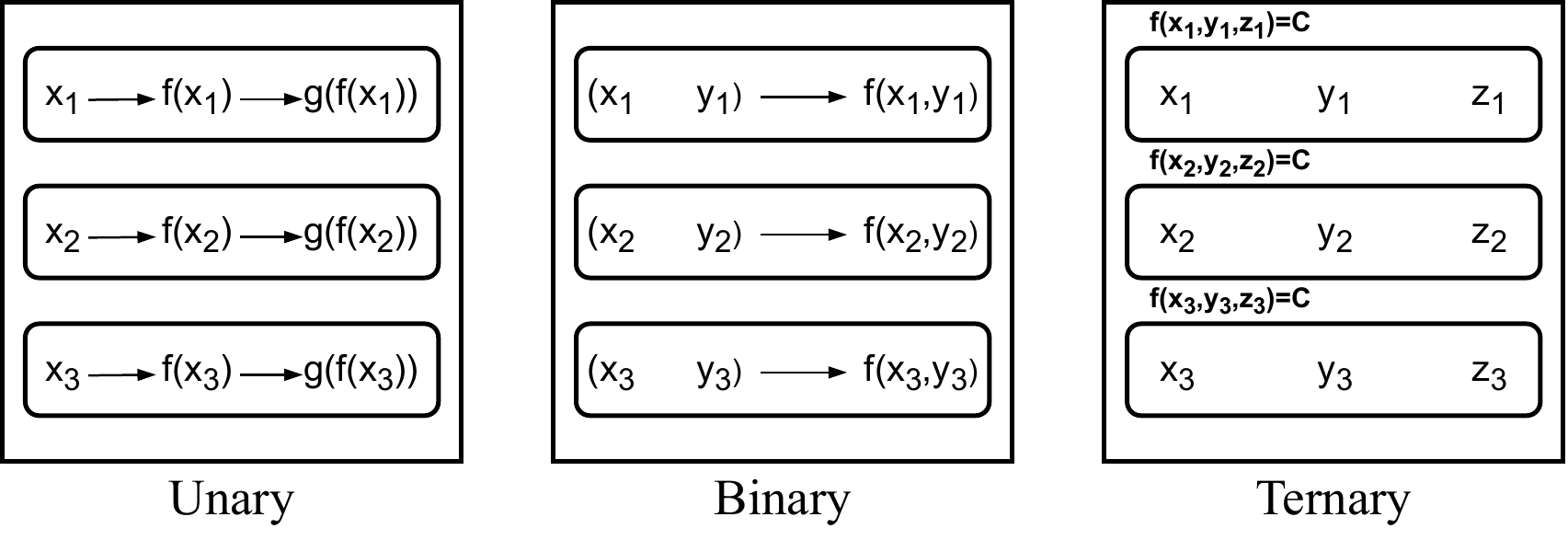}
    \caption{Three categories of relations commonly used in 3$\times$3 RPM problems \citep{wang2015automatic}.}
    \label{fig:3-relations}
\end{figure}

In particular, an variation rule is represented as an instantiation of Equation \eqref{eq-variation-pattern} and \eqref{eq-constraint}, 
{\begin{align}
    \exists \alpha \: \: \forall i \in \{1, 2, 3\}  \: \: \exists o_{i1}, o_{i2}, o_{i3} \: \: P (\alpha, o_{i1}, o_{i2}, o_{i3}) \label{eq-variation-pattern} \\
    P (\alpha, o_{i1}, o_{i2}, o_{i3}) = Unary (\tau (o_{i1}, \alpha), \tau (o_{i2}, \alpha), \tau (o_{i3}, \alpha)) \wedge \label{eq-constraint} \\
    Binary (\tau (o_{i1}, \alpha), \tau (o_{i2}, \alpha), \tau (o_{i3}, \alpha)) \wedge \nonumber \\
    Ternary (\tau (o_{i1}, \alpha), \tau (o_{i2}, \alpha), \tau (o_{i3}, \alpha)) \nonumber
\end{align}}
where $\alpha$ is an attribute, $i$ is the row index, $o_{ij}$ is a geometric elements in the figure of Row $i$ and Column $j$, $\tau (\alpha, o_{ij})$ is the value of $\alpha$ of $o_{ij}$, and $P$ is a predicate that describes the variation pattern of attribute $\alpha$ in each row. In Equation \eqref{eq-constraint}, the predicate $P$ further equals a conjunction of three predicates --- $Unary$, $Binary$, and $Ternary$ --- representing three categories of relations commonly used in 3$\times$3 RPM problems, as illustrated in Figure \ref{fig:3-relations}.

An interesting observation of Figure \ref{fig:3-relations} is that, mathematically, the unary relation is a proper subset of the binary relation, which is a proper subset of the ternary relation. That is, the ternary relation is thus mathematically sufficient to generate all the items. However, interpreting the same variation as unary, binary and ternary relations probably requires different working memory abilities and goal management abilities.
Therefore, these three categories are probably cognitively ``disjoint'', and need to be separately included in a generator program to achieve a better control over psychometric properties.

Equation \eqref{eq-variation-pattern} and \eqref{eq-constraint} represent only the variation pattern of a single attribute $\alpha$. There could be multiple variation patterns of different attributes in a matrix.  
It is possible that some attributes are not assigned instantiations of Equation \eqref{eq-variation-pattern} and \eqref{eq-constraint}, and they could be given either constant values or random values across figures. Random values may cause distracting effects in generated items, which is similar to the nonharmonic perceptual organizations in \citep{primi2001complexity}.

Human data on the generated (non-distracting) items showed that the overall difficulty and rule-wise difficulty (number of rules) were similar to the items in Advanced Raven's Progressive Matrices (APM). However, as the author pointed out, their generator could not synthesize all the items in APM for some underlying transformations were hard to implement. When the items were created with distracting attributes, the generated items became much more difficult for human subjects.


\subsection{IMak Package --- Open Source}

\begin{figure}[h]
    \centering
    \begin{subfigure}{0.32\textwidth}
        \centering
        \includegraphics[width=\textwidth]{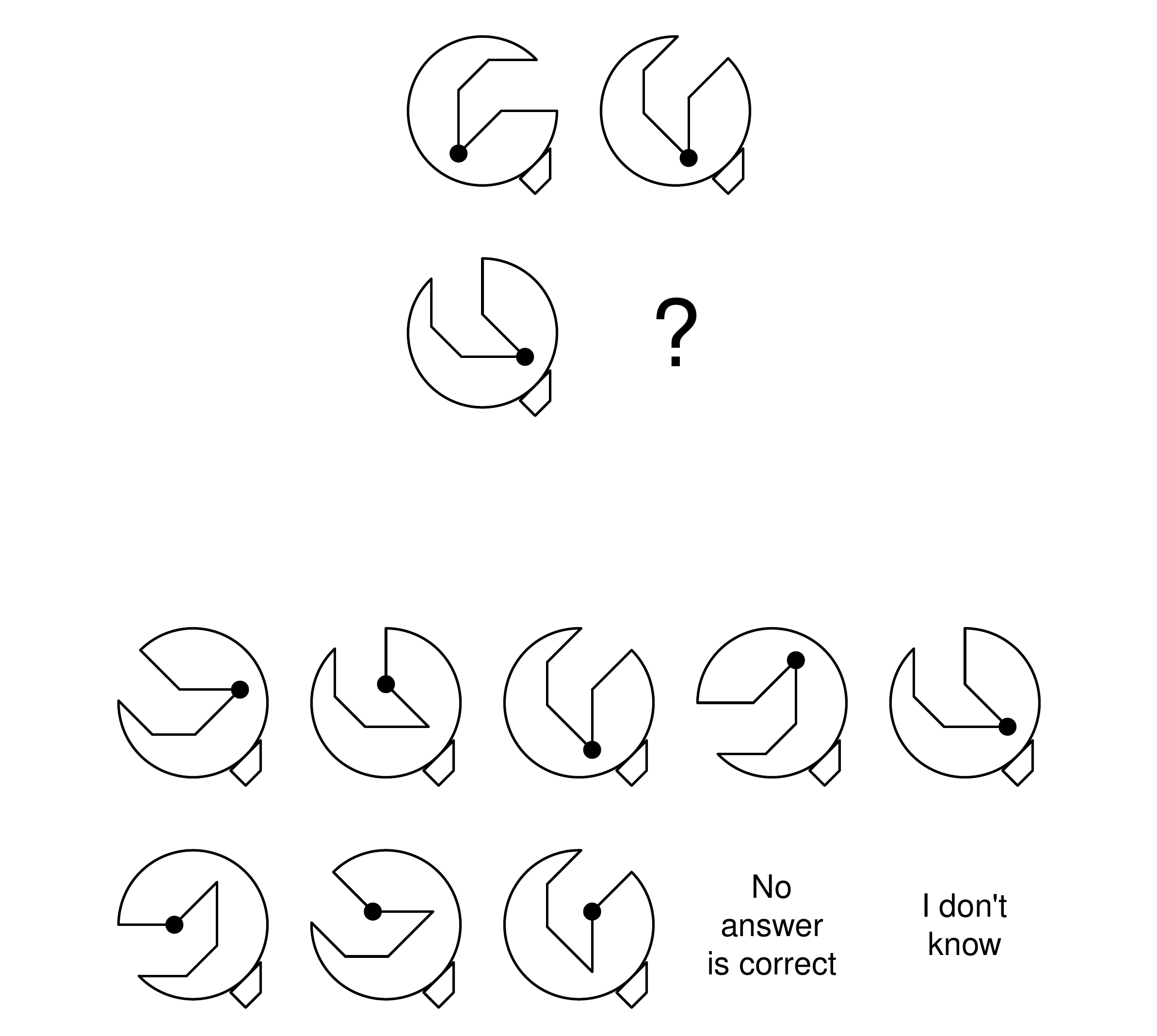} 
        \caption{Main shape rotation}
        \label{fig:imak_main_rot}
    \end{subfigure}
    \begin{subfigure}{0.32\textwidth}
        \centering
        \includegraphics[width=\textwidth]{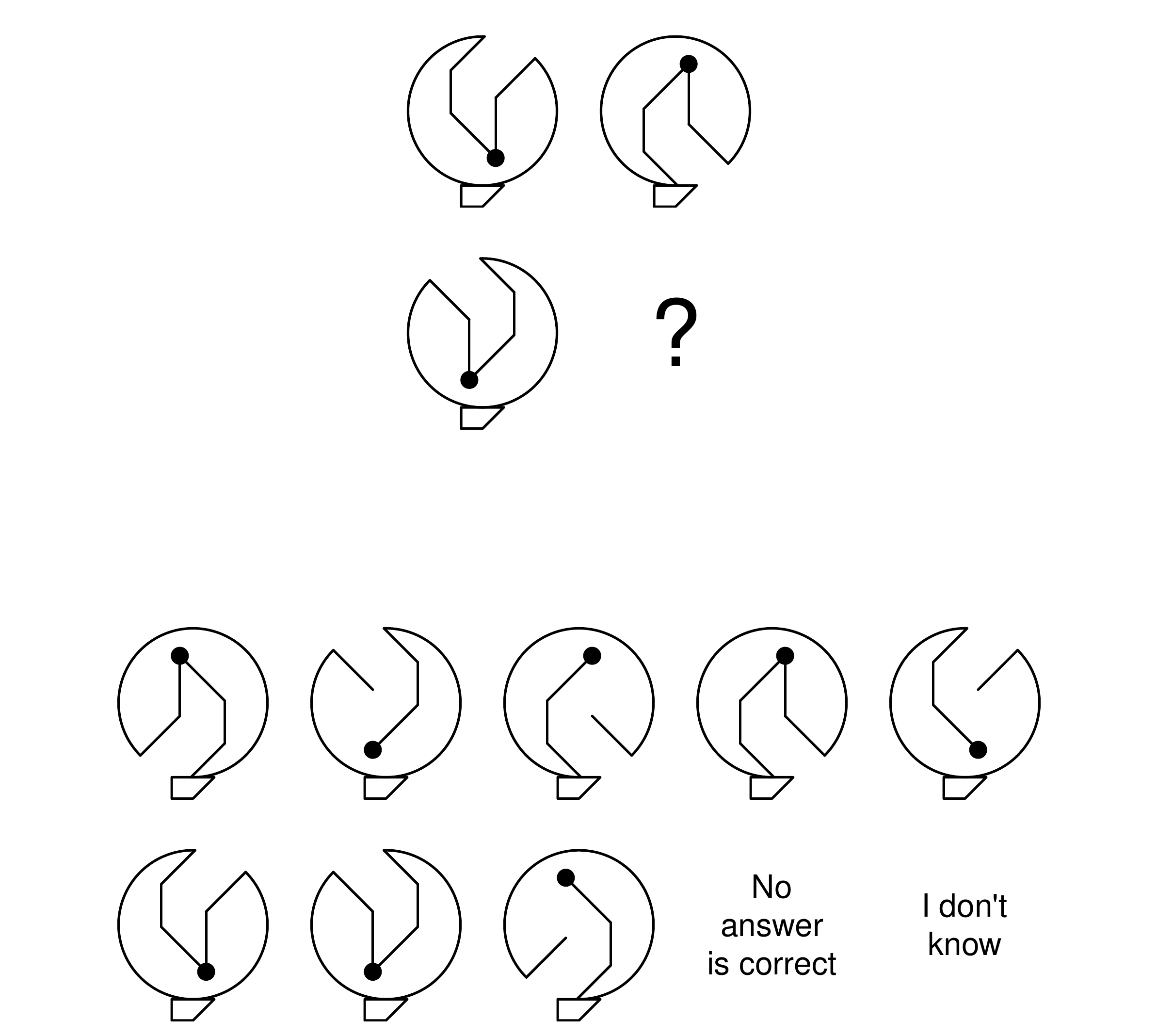}
        \caption{Main shape reflection}
        \label{fig:imak_mirror}
    \end{subfigure}
    \begin{subfigure}{0.32\textwidth}
        \centering
        \includegraphics[width=\textwidth]{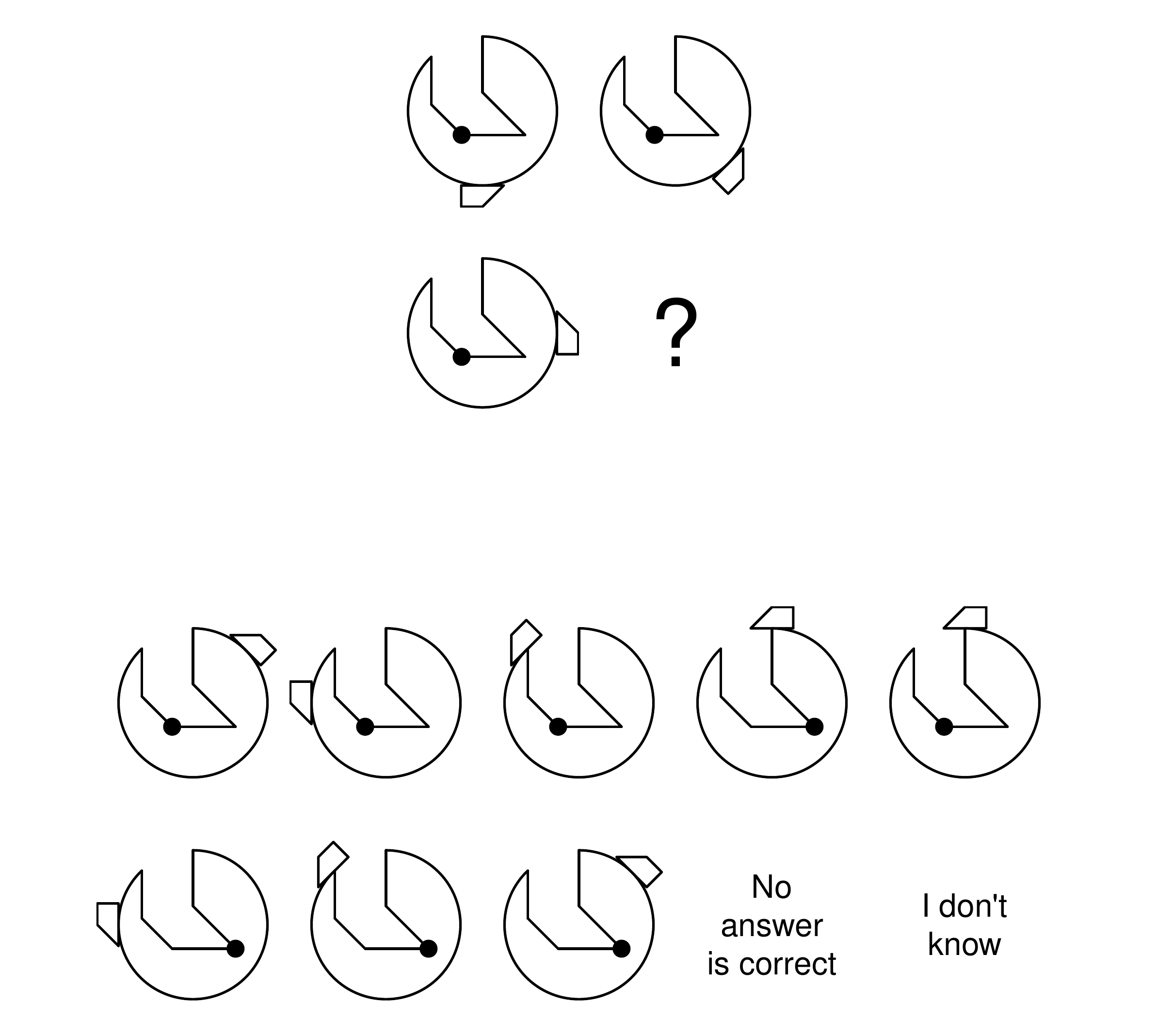}
        \caption{Trapezium rotation}
        \label{fig:imak_trap}
    \end{subfigure}
    
    \begin{subfigure}{0.32\textwidth}
        \centering
        \includegraphics[width=\textwidth]{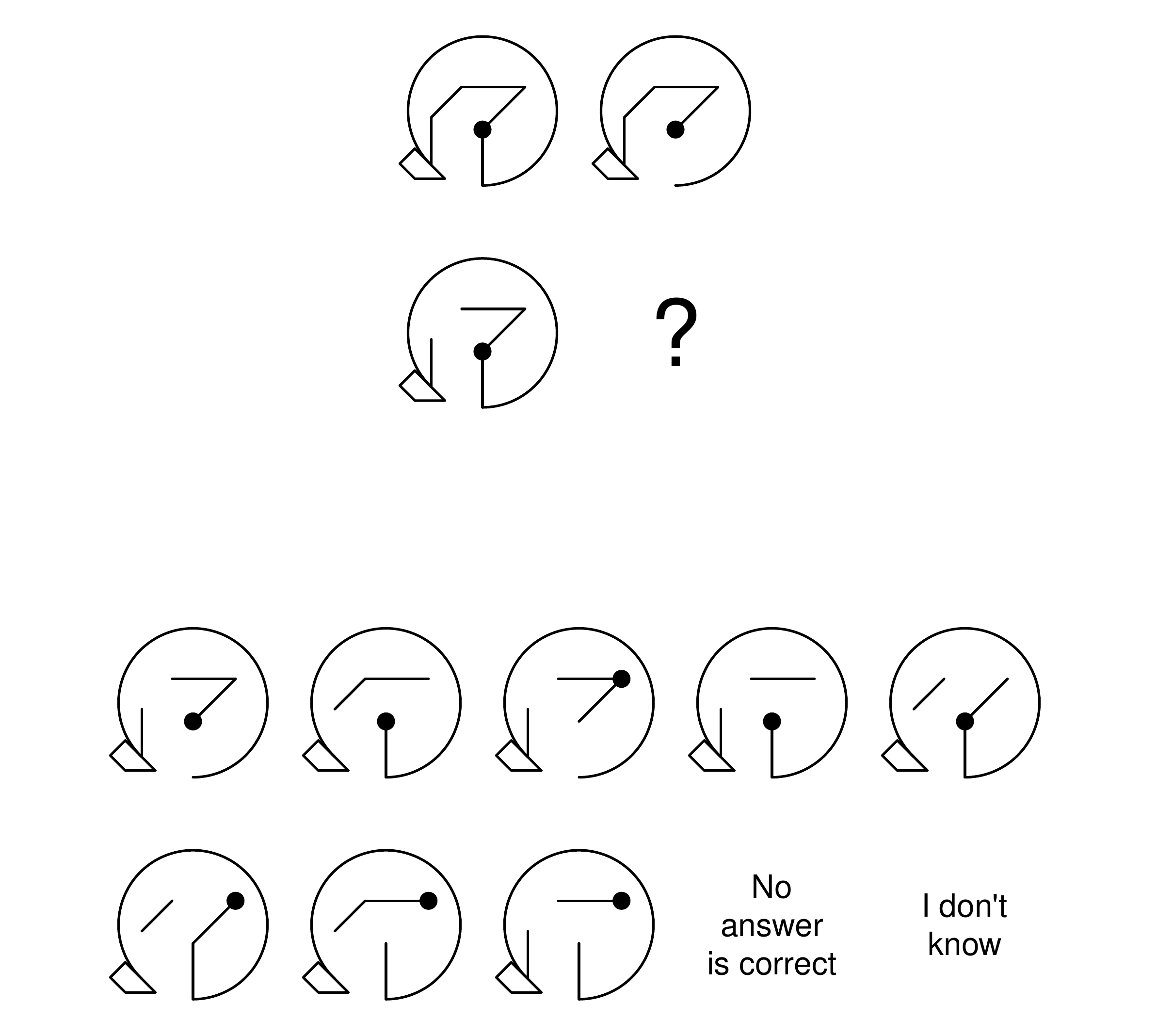}
        \caption{Line segment subtraction}
        \label{fig:imak_subtract}
    \end{subfigure}
    \begin{subfigure}{0.32\textwidth}
        \centering
        \includegraphics[width=\textwidth]{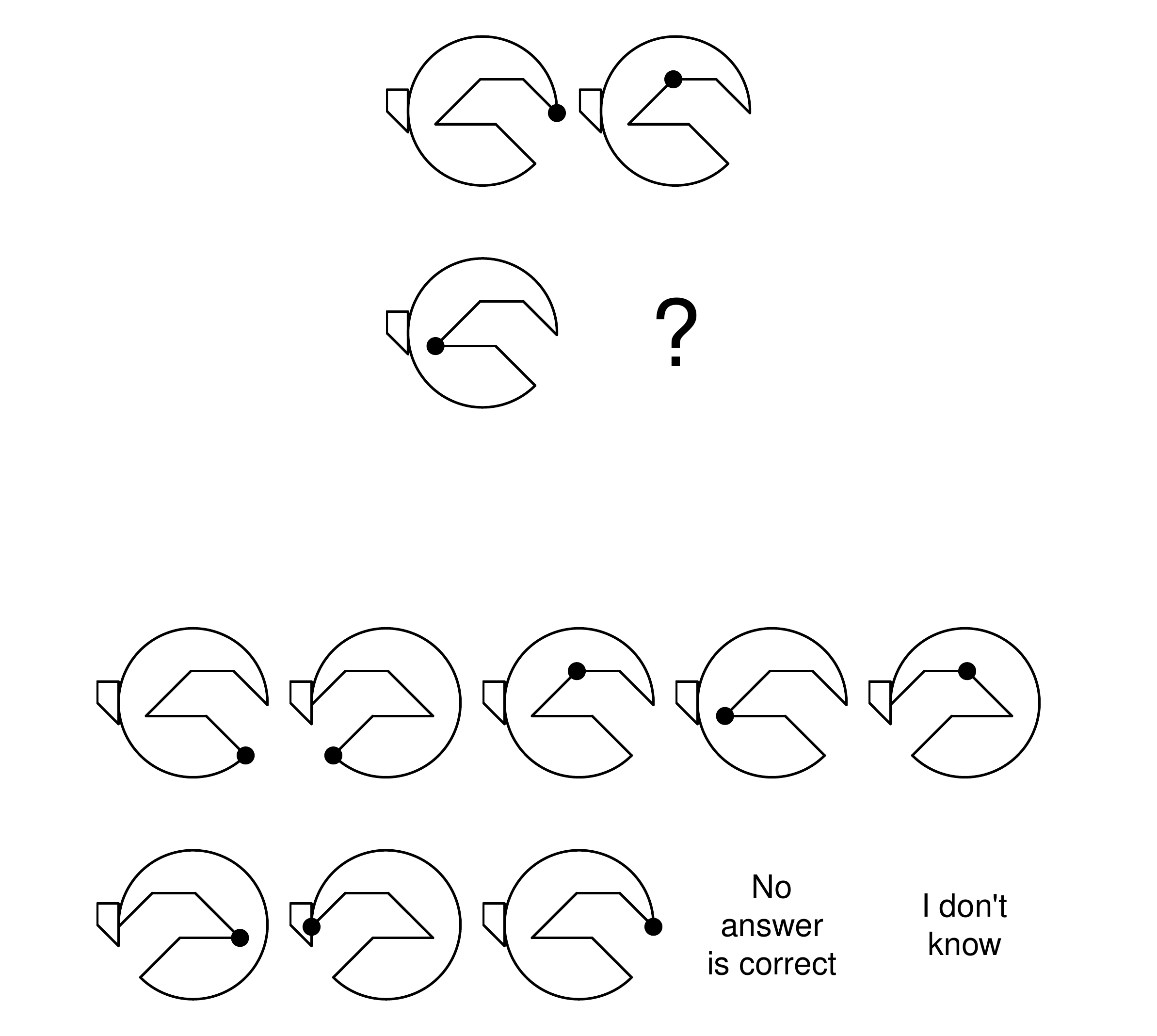}
        \caption{Dot movement}
        \label{fig:imak_dot_mov}
    \end{subfigure}

\caption{Example items generated by the IMak package. Each item exemplifies a single basic rule. The correct answer for each item is arranged as the first option for demonstration purposes.}
\label{fig:imak}
\end{figure}

\citet{blum2018automatic} released IMak --- an open source R package --- to study how different types of variation rules could affect item difficulty. The generator was designed to manipulate the types of rules while keeping other factors constant, and the generated items look quite different from what one would expect in ordinary FAPs, such as RPM items. For example, Figure \ref{fig:imak} shows example items that we created through the package, each of which exemplifies a basic rule type. With the current release (version 2.0.1), the geometric elements are limited to the main shape (the broken circle plus the polyline in it), the trapezium that is tangent to the main shape, and the dot at one of the corners of the polyline. Furthermore, the size and shape of each element are fixed for all generated items, but the position, orientation and existence would vary in finite domains according to 5 basic rules. 

IMak has four other distinctive features. Firstly, IMak filled the vacancy of 2$\times$2 format of AIG. Being affected by the famous work of \citep{carpenter1990one} on RPM, the vast majority of AIG works would only consider the 3$\times$3 format. However, the original RPM test also contains 2$\times$2 items, and the proportional geometric analogies (see the 2$+$2 format in Figure \ref{fig:22gap}) could also be put into the 2$\times$2 format. Secondly, the answer set contains two more meta-options ``no correct answer'' and ``I don't know'', which encourage subjects to solve the items more constructively rather than eliminating responses. Thirdly, the variation of one element could depend on the variation of another element. For example, the dot's moves depend on the variation of the main shape, since the dot only moves along the polyline in the main shape. This kind of variation is very rare in item writing of FAPs, and it represents an extra level of complexity of FAP items.

Last but not least, IMak used a rule-dependent strategy to generate incorrect answer choices. For 1-rule items, 4 distinct values of the attribute of the rule are sampled, including the correct value; since all other attributes remain constant in the matrix, another random attribute is chosen and sampled for 2 values. The resulting 8 (4$\times$2) combinations make the 8 options in the answer set. For 2-rule items, 3 values are sampled for each of the 2 attributes of the 2 rules, resulting in 9 combinations, and one of them is discarded. For 3-rule items, 2$\times$2$\times$2 combinations are sampled. For four-rule items, 2$\times$2$\times$2$\times$2 combinations were sampled, and half of them are discarded.

In a human experiment, 23 generated items were administered to 307 participants from Germany, Indonesia, and Argentina.
Reliability, validity and unidimensionality were initially verified by the experiment results. Particularly, item difficulty could be \textbf{partly} predicted from the number and type of rules based on psychometric models, but far from enough for psychometric applications.


\section{Automatically Generating FAPs for Fitting Data-Driven AI Models}




\subsection{Procedurally Generated Matrices}
Based on the five rules in \citep{carpenter1990one}, \citet{barrett2018measuring} continued the first-order logic approach of \citet{wang2015automatic} and created a large-scale (1.2M) dataset of FAPs --- Procedurally Generated Matrices (PGM). Since the generator program is not made public, our discussion is solely based on the description in  \citep{barrett2018measuring} and our observation of the dataset.

\begin{figure}[ht]
    \centering
    \includegraphics[width=\textwidth]{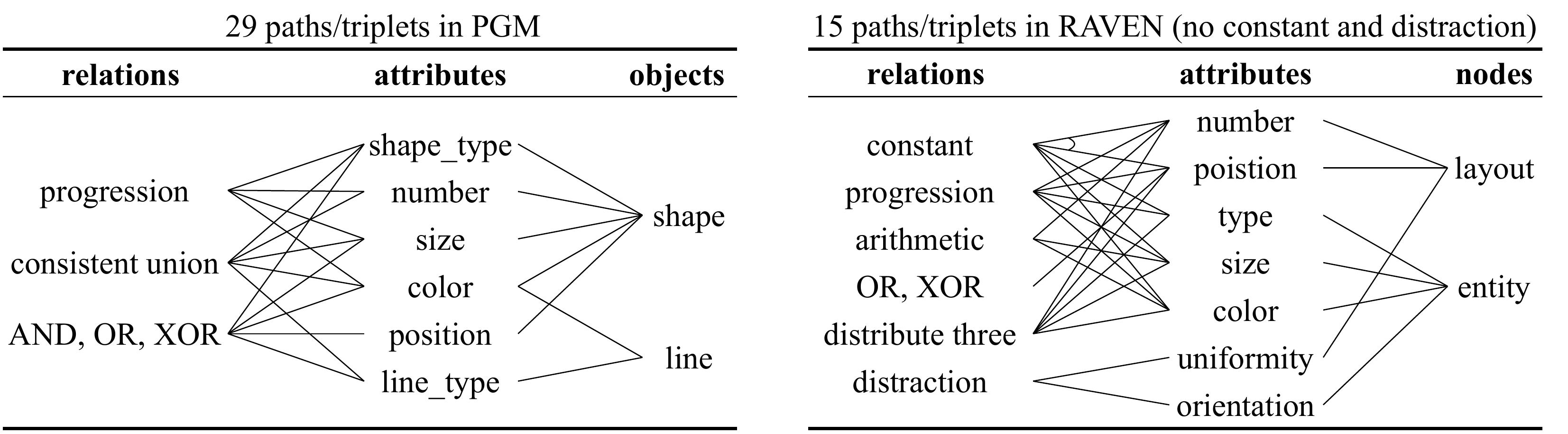}
    \caption{Left:  The dependencies among relations, objects, and attributes used to generate the Procedurally Generated Matrices (PGM) dataset \citep{barrett2018measuring}. Each path corresponds to a $[r, o, a]$ triplet representing a possible type of variation pattern in the matrices. As one can check, there are 29 paths, i.e. $[r, o, a]$ triplets, in the graph. Note that in \citet{barrett2018measuring} did not differentiate between ``shape\_type'' and ``line\_type'' and referred to both of them as ``type''. In their implementation, these two attributes are treated as two distinct ones.  Right: The dependencies among relations, nodes, and attributes were used to generate the RAVEN dataset.  Note that we listed ``distraction'' as a rule in this graph to indicate that uniformity and orientation are distracting attributes. The two paths from constant through number and position to layout are treated as a single rule in RAVEN. Therefore, there are 15 paths, but 14 rules, in the graph.}
    \label{fig:roa}
\end{figure}

In PGM, an instantiation of Equation \eqref{eq-variation-pattern} and \eqref{eq-constraint} in the first-order logic approach was denoted by a triplet $[r, o, a]$ of relation, object and attribute. Object here means the geometric element in previous sections of AIG for human intelligence tests. But since most reviewed works in this section use object instead of element, we follow this convention in this section. These three factors are not independent, and we visualized their dependencies in Figure \ref{fig:roa}, which contains 29 paths from the left to the right, corresponding to 29 $[r, o, a]$ triplets. This number equals the number of triplets mentioned in \citep{barrett2018measuring} (but it did not provide a list of the 29 triplets). 

As shown in Figure \ref{fig:roa}, the objects in PGM are classified into two disjoint subsets --- shape and line. In the shape subset, closed shapes are arranged in 3$\times$3 grid (fixed positions in this case) in each figure. In the line subset, line drawings, spanning the whole figure, are always centered in each figure. There exist items generated by superimposing a shape item on a line item, but these two items are completely independent in terms of variations. Thus, in Table \ref{tab:tech-details}, we split PGM into two rows to describe it more clearly.

The generation procedure of a PGM item could be described by 5 steps: (a) sample 1 to 4 triplets from the 29 triplets described in Figure \ref{fig:roa}. 
(b) determine the analogical direction for each triplet: row or column; (c) sample attribute values for each triplet from their domains. 
(d) determine the attribute values for unspecified attributes (either constant or random). 
and (e) render all attribute values into a pixel image of the matrix.

\subsection{Relational and Analogical Visual rEasoNing}

\begin{figure}[ht]
    \centering
    \includegraphics[width=\textwidth]{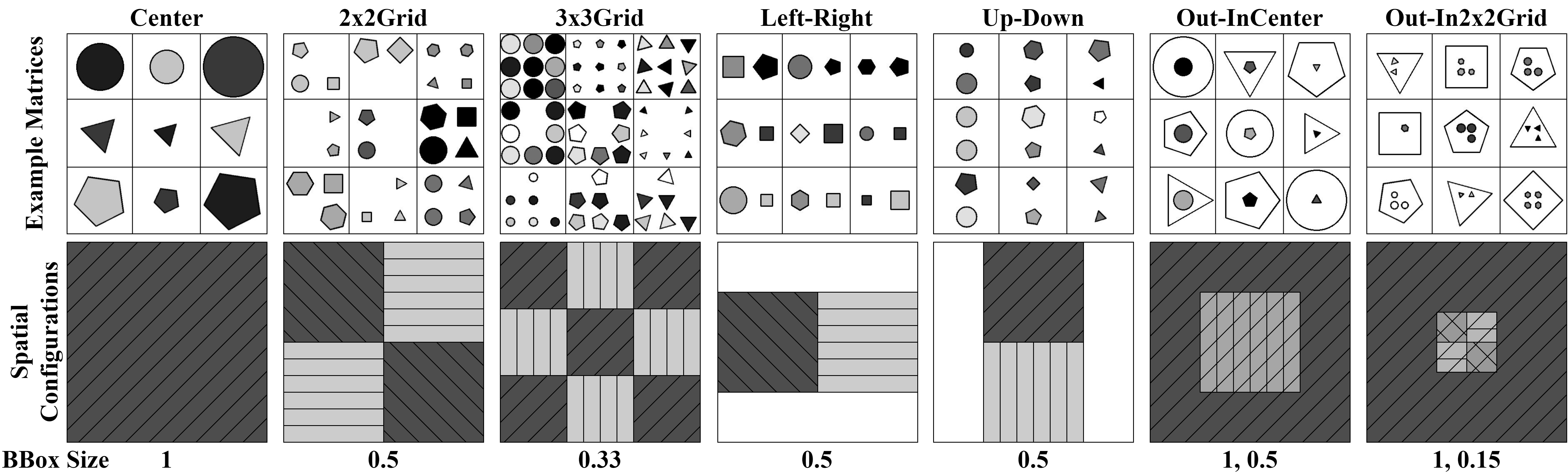}
    \caption{7 hardcoded spatial configurations --- center, 2x2Grid, 3x3Grid, Left-Right, Up-Down, Out-InCenter, and Out-In2x2Grid --- are used to arrange objects in each figure in the RAVEN dataset. Each configuration is represented by the bounding boxes that objects could occupy. The position and size of each bounding box are hardcoded in the generator program. An example matrix for each configuration is given in the first row. Note that not every bounding box has to be occupied, but every object has to be in one of the bounding boxes.}
    \label{fig:7-configs}
\end{figure}

Due to the limited expressive power of $[r, o, a]$ in PGM, the spatial configuration, as an important dimension of perception organization, is highly restricted --- 3$\times$3 grid for the shape subset, all-centered for the line subset, and superimposing a shape item on a line item for the intersection of these two subsets. \citet{zhang2019raven} attempted to encode the spatial configuration and generated the Relational and Analogical Visual rEasoNing (RAVEN) dataset, which included 7 hardcoded spatial configurations, as shown in Figure \ref{fig:7-configs}. The source code of RAVEN's generator is available online, inspection of which drives our discussion.  

\begin{figure}[ht]
    \centering
    \includegraphics[width=\textwidth]{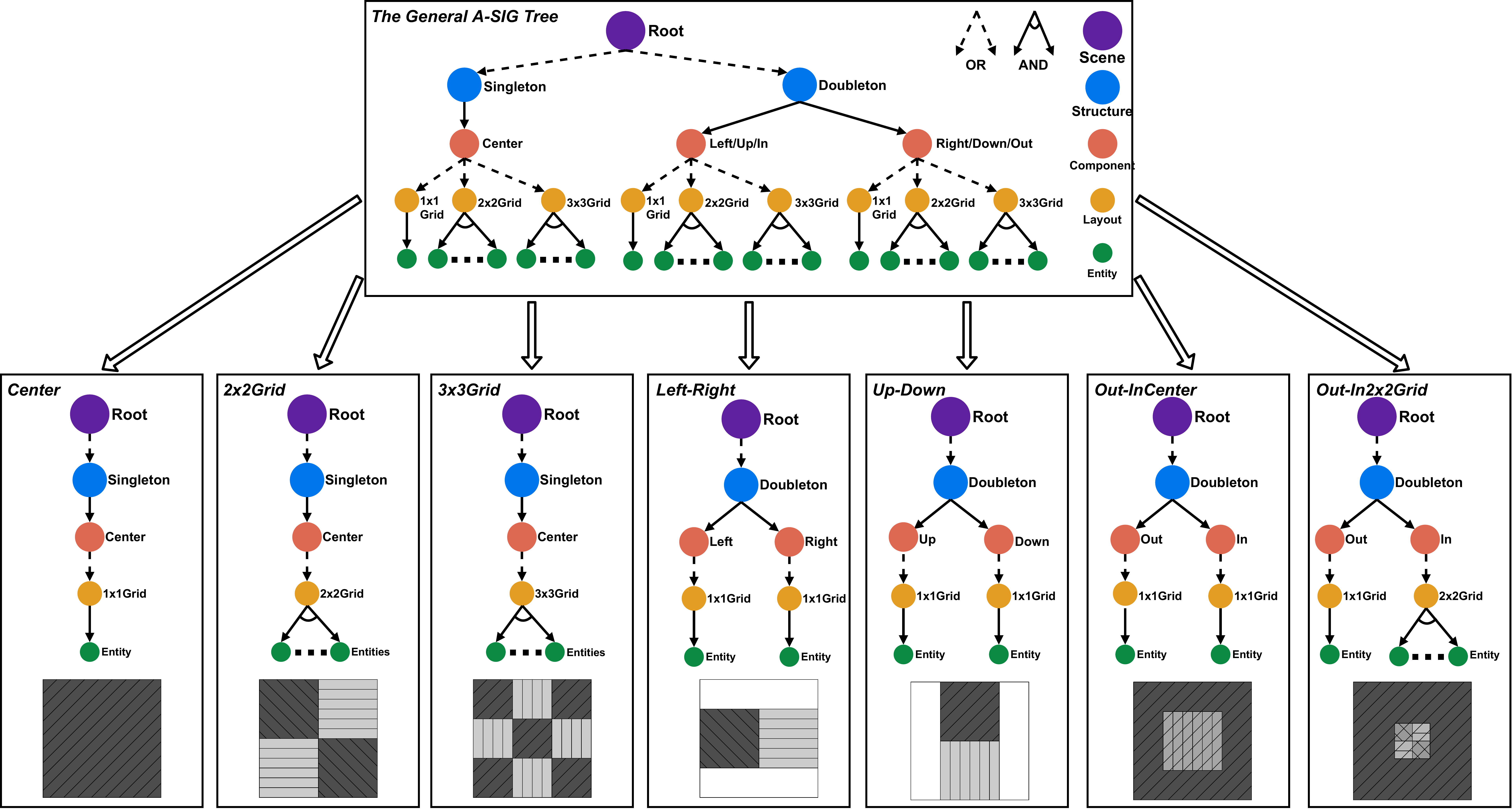}
    \caption{The general A-SIG tree and 7 specific A-SIG trees were used in the RAVEN dataset (image adapted from \citep{zhang2019raven} by adding technical details from the source code of the generator program). The root node denotes the scene that the images of interest would describe. The structure nodes are the containers of different spatial structures. A structure is composed of components that could be overlaid with each other. Each component has its own layout and, more importantly, variation rules, which are independent of other components. The layout node, as its name indicated, contains the attributes specifying the number and positions of geometric objects. Entities represent geometric objects whose attributes are invariant to number and position.}
    \label{fig:A-SIGs-RAVEN}
\end{figure}

The 7 configurations are derived from a more general symbolic representation framework of images --- Attributed Stochastic Image Grammar (A-SIG). In A-SIG, a potential image is described by a tree structure, where the conceptual granularity becomes finer and finer toward the leaf nodes. To generate RAVEN, the tree structure is predefined as a general A-SIG tree as shown in Figure \ref{fig:A-SIGs-RAVEN}, 
which consists of 5 conceptual levels --- scene, structure, component, layout, and entity --- and uses a stochastic tree-traversal process to generate items. 
However, the 7 configurations in RAVEN were hardcoded in the language of A-SIG, rather than generated through this stochastic traversing process, which could have made RAVEN more diverse in spatial configuration, and, meanwhile, more challenging.

\begin{figure}[ht]
    \centering
    \includegraphics[width=\textwidth]{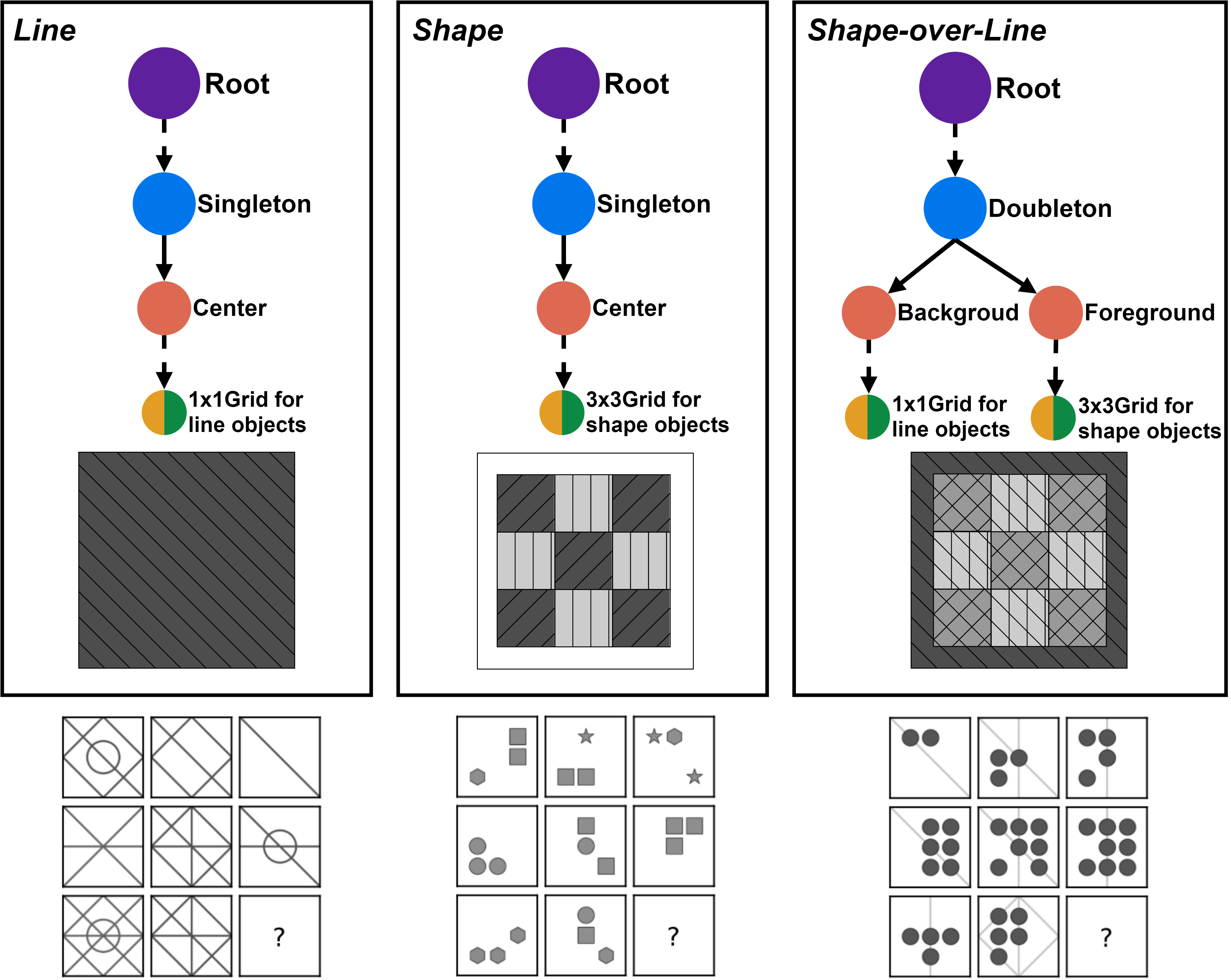}
    \caption{The spatial configurations of PGM could also be represented in A-SIG to compare with RAVEN. Based on observations, there are 3 spatial configurations in PGM --- line, shape, and shape-over-line. The example matrix is given for each configuration at the bottom.}
    \label{fig:A-SIGs-PGM}
\end{figure}

To compare with the PGM dataset, we represent PGM items also in A-SIG, as shown in Figure \ref{fig:A-SIGs-PGM}. The line configuration of PGM is basically the same as the center configuration of RAVEN except that the entity types (shape) are different. The shape configuration of PGM is almost the same as the 3x3Grid configuration of RAVEN except that bounding box sizes are slightly different. The shape-over-line configuration of PGM is also conceptually similar to the double-component configurations of RAVEN. The general difference between PGM and RAVEN lies in the layout and entity nodes. As shown in Figure \ref{fig:A-SIGs-PGM}, the PGM dataset is not able to separate the concepts of entity and entity layout due to the limited expressive power of triplets $[r, o, a]$. That is, the object $o$ takes the roles of both layout and entity nodes, but could not play the roles effectively and simultaneously.

Unlike PGM, RAVEN inherited all the five rules from \citep{carpenter1990one}. Moreover, the ``addition-and-subtraction'' rule type in RAVEN contains not only figure addition and subtraction, such as the set operations ``OR and XOR'', but also arithmetic addition and subtraction, which were not in \citep{carpenter1990one}. Since these two operations are conceptually different, we refer to the arithmetic addition and subtraction as ``arithmetic'', and the figure addition and subtraction as ``OR and XOR''. In addition, the ``distribution-of-three-values and distribution-of-two-values'' in \citep{carpenter1990one} were merged into a single rule by considering the latter as a special case of the former with a null value for one of the three values. Therefore, RAVEN has a slightly different rule set compared to PGM. Similarly, we could represent the variation rules of RAVEN also as triplets --- $[r, n, a]$ where $n$ represents nodes (layout or entity) in A-SIG trees, and $r$ and $a$ are relations and attributes, being the same as PGM. Then Figure \ref{fig:roa} shows the dependencies among $r$, $n$ and $a$. 


PGM and RAVEN are similar in some respect, and meanwhile, different in others. In particular, they share two similarities. First, their choices of attributes, attribute domains, and rule types are similar. For example, they both forbid number-rule and position-rule from co-occurring in an item because these two attributes would probably conflict with each other. Second, although RAVEN has more spatial configurations, these configurations are not structurally different from PGM (as can be seen from the comparison of their A-SIG trees). 


Meanwhile, PGM and RAVEN are different in two respects. First, they are different in the number of rules in an item. In PGM, 1 to 4 triplets were sampled from the 29 triplets.
In contrast, in a RAVEN item, every attribute is governed by a rule except the two distracting attributes (uniformity and orientation). Thus, there are 4 rules (for number/position, type, size, and color, respectively) in each RAVEN item. Second, the rules in RAVEN are all row-wise while the rules in PGM are either row-wise or column-wise.

\subsection{Context-Blind Issue}

The answer sets in RAVEN were generated in the same way as the first-order logic approach. Each incorrect answer was obtained by modifying a single attribute of the correct answer. 
The method kept the maximum level of distracting and confusing effect of incorrect answer choices, because one has to identify all the variation rules to solve the problem. Meanwhile, ignoring any rule would lead to multiple choices. However, this design has a major drawback for failing the context-blind test. In a RAVEN item, the incomplete matrix could be considered as the context that provides information for solving the problem. Failing the context-blind test means that it is possible for human subjects or computational models to select the correct answer while turning blind to the context.

\citet{hu2021stratified} and \citet{benny2021scale} pointed out that RAVEN failed the context-blind test. They provided evidence that computational models could achieve high accuracies (from 70\%+ to 90\%+) by training only on the answer sets of RAVEN. In addition, the context-blind accuracies were even better than the accuracies that could be achieved by training the same data-driven model on complete items. This implies that data-driven AI models are capable of capturing the statistical regularities in the answer sets. The reason for this context-blind issue obviously lies in the answer set generation, in which every incorrect answer choice is a variant by modifying a single attribute of the correct answer choice. Therefore, the correct answer is the answer choice that possesses every common feature among all the choices (or, equivalently, the one most similar to every other choice).

\begin{figure}[ht]
    \centering
    \includegraphics[width=0.7\textwidth]{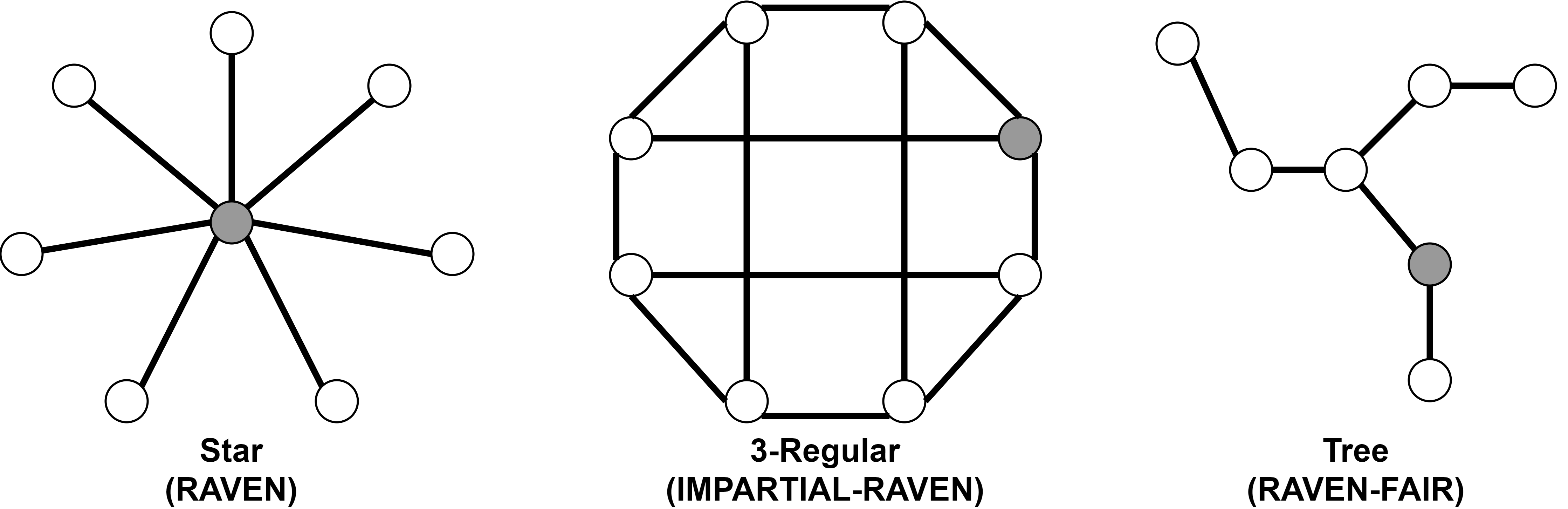}
    \caption{The answer sets of three versions of the RAVEN dataset depicted in graphs. Each vertex is an answer choice and two adjacent vertices differ by one attribute.}
    \label{fig:3-RAVEN-answer-generation}
\end{figure}

Both \citet{hu2021stratified} and \citet{benny2021scale} proposed their own solutions to this issue --- the Impartial-RAVEN and RAVEN-FAIR datasets. These two datasets have the same context matrices as the original RAVEN and regenerated the answer sets in different ways. Interestingly, the similarity and difference among these three versions of answer sets could be clearly illustrated by putting them in graphs (simple graph, not digraph or multigraph). If we represent each answer choice as a vertex and each modification of a single attribute as an edge, then the answer sets of the three versions could be depicted by the graphs in Figure \ref{fig:3-RAVEN-answer-generation}. The answer set of the original RAVEN is created by modifying an attribute of the correct answer. Thus, its graph is a star centered at the correct answer (the solid vertex). 
And what the aforementioned computational models captured was the unique center of this star graph.

\citet{hu2021stratified} proposed the Impartial-RAVEN whose each answer set could be represented by a 3-regular graph in Figure \ref{fig:3-RAVEN-answer-generation}. 
\citet{benny2021scale} proposed another less regulated way to generate answer sets. Starting from an initial answer set consisting of only the correct answer, an answer choice is randomly selected from the current answer set, and then an attribute of the selected answer choice is randomly altered; this process repeats, resulting in a random tree-like structure as shown 
on the right in Figure \ref{fig:3-RAVEN-answer-generation}. 

These two enhanced versions of RAVEN were tested by context-blindly training the benchmark model in \citep{zhang2019raven} and the CoPINet model in \citep{zhang2019learning}. The performance decreased to below 20\%. Ideally, a human subject (or a computational model) who context-blindly works on the RAVEN items should perform at most as good as a random guess, i.e. $1/8=12.5\%$, which implies that the answer set per se does not provide any useful information for solving the item. However, in the practice of item writing, to maintain a certain level of distracting and confusing effect of incorrect answer choices, the majority of incorrect answer choices must share some similarities with the correct answer. Otherwise, it would be relatively easy for subjects to find the correct answer, because incorrect answers would be perceptually distinct from the context. Therefore, a reasonable context-blind performance would be slightly higher than the random guess, and determined by the researcher's judgment.


An interesting observation on the difference between these two enhancements could be made by comparing their graphs in \ref{fig:3-RAVEN-answer-generation}. If we consider a single trial (in a probabilistic sense) where we context-blindly give a subject (or an AI model) an item from Impartial-RAVEN and an item from RAVEN-FAIR, the probability that this subject solves the Impartial-RAVEN item would not be significantly different from that of solving the RAVEN-FAIR item. However, if we repeat this trial (with different items) again and again, the performance on RAVEN-FAIR would probably gradually exceed the performance on Impartial-RAVEN, of course, assuming that the subject is intelligent enough and able to figure out the graph structures, and thus makes an educated guess by selecting the ``center'' (or the max-degree vertex) of trees in a probabilistic sense. In this case, we would say that the RAVEN-FAIR is context-blind valid at the item level, but not at the dataset level.


\section{Discussion and Conclusion}

In this paper, we reviewed automatic item generation (AIG) of figural analogy problems (FAPs). We classified the works into two groups by their purposes --- whether it is for human ability tests or for fitting and evaluating AI models. The works in the first group aim at not only generating items but also good psychometric properties. 
The works in the second group could be seen as the continuation of the works in the first group, but the psychometric aspects were usually not stressed. 

For example, 18 PGM items were administered to human participants. Participants without prior experience of FAPs would often fail almost all the items, whereas participants with prior experience could score up to 80\%. Therefore, the discriminative power is too large and the ability range that could be effectively tested is too narrow. This also implies that the datasets used to evaluate data-driven AI models are not necessarily suitable to evaluate human cognitive ability. 

More importantly, this gives rise to another interesting question --- how do we assess the analogical ability of AI models trained and evaluated on the large datasets in the second group? On one hand, some data-driven AI models indeed performed well on AIG items that posed great challenges to human subjects; on the other hand, training on the large-scale datasets specially prepared the AI models for a highly restricted subset of the problem domain, but human subjects, who were not trained at all, or just trained on several representative exemplars from this subset, could perform well and even generalize well in the problem domain.

This is not the first time that this question is asked, for example, \citep{detterman2011challenge}. Efforts have been made to address this issue by many researchers. \citep{bringsjord2003artificial, bringsjord2011psychometric} tried to address this issue by incorporating the implementation and evaluation of AI models into a general concept of psychometric AI. By reviewing roughly 30 computational models for solving intelligence tests, \citet{hernandez2016computer} proposed that (a), instead of collecting items, we should collect item generators, and (b) the generated items should be administered to machine and human (and other animals) alike (universal psychometrics). All these propositions are constructive and, meanwhile, suggest much higher requirements for AIG studies.

In the domain of FAPs, which is loosely (even vaguely) defined, a practical roadmap for future AIG studies could better start from studying the problem domain per se instead of continuing to construct ad hoc generator programs. Huge uncharted territories lie in the complexity factors such as the types of elements and rules and perceptual organization. Current AIG item banks and datasets are far below the level of the flexibility and diversity that human item writers can achieve. For example, the spatial configurations in PGM and RAVEN are hardcoded and basically stack independent items together; inter-element variation, in which the variation of one element depends on the variation of another element, is also very rare, as are perceptually and conceptually ambiguous analogies.  

At last, our review is by no means comprehensive and exhaustive; and a combined interdisciplinary effort is definitely a desideratum for future research.

\newpage
{\parindent -10pt\leftskip 10pt\noindent
\bibliographystyle{cogsysapa}
\bibliography{references}}


\end{document}